 \documentclass[final,5p,times,twocolumn]{elsarticle}

\usepackage{amssymb}
\usepackage{graphicx}
\usepackage{amsmath}
\usepackage{rotating}
\usepackage{float}
\usepackage{caption}
\usepackage{tabularray}
\usepackage{natbib} 
\usepackage{placeins}
\usepackage{enumitem} 
\usepackage{algorithm}
\usepackage{algpseudocode}

\journal{Energy}

\begin{document}

\begin{frontmatter}


\title{Step-adaptive multimodal fusion network with multi-scale cloud feature learning for ultra-short-term solar irradiance forecasting}


\author{Jingxin Zhang, Xiaoqin Wang} 

\affiliation{organization={School of Automation, Southeast University},
            city={Nanjing},
            postcode={210096}, 
            country={China}}

\begin{abstract}
Ultra-short-term solar irradiance prediction is critical for photovoltaic system dispatch and power grid stability. Existing approaches suffer from three key shortcomings: single time-series models cannot capture the spatial dynamics of clouds under complex conditions, standard convolutions inadequately represent multi-scale cloud features, and fixed low-frequency compensation strategies fail to adapt to different prediction steps. To address these issues, this proposes a multi-source data fusion model for ultra-short-term irradiance prediction. The model first employs InceptionNeXt to extract multi-scale, multi-directional spatial features from ground-based cloud images. A step-adaptive low-frequency compensation unit is then introduced to dynamically modulate global low-frequency information based on the prediction step. Eventually, the enhanced image features are combined with meteorological time-series features, and a TempAttnLSTM network captures global temporal dependencies for multi-step prediction. Experiments on the public NREL dataset and practical photovoltaic stations in Shandong illustrate the effectiveness of the proposed method compared with several state-of-the-art approaches.

\end{abstract}




\end{frontmatter}

\section{Introduction}\label{sec1}

Growing installed photovoltaic capacity and grid penetration have made intermittent and volatile PV power generation a critical challenge for grid safety and power trading. Reliable solar irradiance forecasting is therefore essential to maintain grid stability and boost renewable energy accommodation \cite{rafati2021high}. In ultra-short-term prediction, dynamic cloud movement triggers sharp fluctuations in irradiance, frequently leading to significant prediction errors. Conventional time-series models only mine statistical features from meteorological data and fail to capture the spatial migration and shading effects of cloud clusters. Therefore, multimodal methods combining ground-based cloud images and meteorological time-series data have emerged as a mainstream solution to advance ultra-short-term solar irradiance forecasting.

Ground-based cloud images have high spatial and temporal resolution, which intuitively reflects cloud morphology, cloud movement, and real-time solar shading behaviors. Such visual data provides physical and interpretable clues for analyzing solar irradiance variations. Alonso-Montesinos et al. \cite{alonso2015skycamera} estimated solar radiation from sky camera images using digital image processing. With the continuous development of deep learning, various advanced models, including CNNs, recurrent neural networks, and Transformers, have been widely adopted for cloud-image-based irradiance forecasting. These techniques support automatic feature extraction and spatio-temporal modeling of dynamic cloud conditions. Shi et al. \cite{shi2024cloudswinnet} proposed a hybrid CNN-Transformer framework for fine-grained ground-based cloud image segmentation.
Zhen et al. \cite{zhen2020deep} combined CNNs and LSTMs to separately extract spatial and temporal features from cloud images. Feng et al. \cite{feng2022convolutional} employed convolutional structures to process cloud image sequences and enhance the modeling capacity of cloud spatio-temporal dynamics. To capture cloud evolutionary characteristics, Huang et al. \cite{huang20233d} proposed a multi-channel 3D ConvLSTM-CNN model for unified spatio-temporal feature learning. Zang et al. \cite{zang2024improving} constructed a spatio-temporal feature interaction framework to strengthen the inherent correlation between cloud patterns and photovoltaic power outputs.

In terms of data-driven optimization, Nie et al. \cite{nie2024sky} utilized heterogeneous cloud images collected from multiple stations and validated that data fusion and transfer learning can effectively improve model generalization. Jonathan et al. \cite{jonathan2024radiant} designed an attention-enhanced CNN model specialized for sequential cloud image-based irradiance prediction. Xu et al. \cite{xu2024minutely} developed a hybrid LSTM-InformerStack architecture to achieve accurate multi-step minute-scale irradiance forecasting. Dai et al. \cite{dai2023photovoltaic} adopted Vision Transformer to fuse cloud images and meteorological time-series data. Wang et al. \cite{wang2019photovoltaic} combined LSTM and convolutional networks for photovoltaic power forecasting. Shi et al. \cite{shi2025ground} integrated CNNs and Vision Transformers to realize precise classification of ground-based cloud images. Ma et al. \cite{ma2025research} further established a multimodal fusion-based ensemble framework, demonstrating the effectiveness of multi-source collaborative modeling for photovoltaic forecasting.
Overall, deep learning approaches are capable of efficiently extracting high-level cloud features from raw images. More importantly, the hybrid modeling strategy that integrates cloud visual information with irradiance and meteorological time-series data significantly improves the model performance in capturing irradiance fluctuations under complex weather. Currently, this strategy has become the mainstream solution for ultra-short-term photovoltaic power forecasting.

Apart from the ongoing updates to model architectures, contemporary research places growing emphasis on multi-source data collaborative modeling and adaptive multi-step prediction for solar forecasting. Caldas et al. \cite{caldas2019very} combined all-sky images with real-time irradiance data and confirmed the benefits of integrating visual observations and ground-measured time-series data. Ajith et al. \cite{ajith2023deep} conducted a comparative analysis among time-series-based, image-based and hybrid deep learning models. Their research illustrated that hybrid models can simultaneously capture historical irradiance trends and spatial features contained in cloud images. From a computer vision perspective, Paletta et al. \cite{paletta2023advances} summarized prevailing solar prediction approaches and highlighted that merging multi-source data  can  reflect dynamic cloud variations and boost model prediction performance. Hendrikx et al. \cite{hendrikx2024all} developed an LSTM-based forecasting method using all-sky images. Additionally, Ansong et al. \cite{ansong2025very} adopted low-cost all-sky cameras together with a CNN-LSTM framework, further demonstrating the practical potential of cloud image-based deep learning models in short-term photovoltaic forecasting. Dou et al. \cite{dou2025multimodal} further explored multimodal irradiance forecasting with cloud imagery and time-series data.

Although multimodal approaches are efficient for short-term irradiance forecasting, several technical challenges still hinder further performance improvement. Firstly, cloud clusters present complex multi-scale and multi-directional properties. Small broken clouds, large stratus clouds and fast-moving cumulus clouds differ significantly in spatial structure, boundary shape and movement characteristics. Most existing models rely on standard convolution or single-scale feature extraction modules, which cannot adaptively extract cloud features across various scales and directions, thus yielding insufficient feature representation. Secondly, limited by local receptive fields and hierarchical pooling operations, conventional CNNs gradually lose global low-frequency information with the deepening of network layers. Such defects prevent models from capturing the overall evolutionary patterns of clouds, thereby degrading the prediction accuracy for long-step forecasting tasks. Wu et al. \cite{wu2024multiscale} developed the multiscale low-frequency memory network (MLFM) to supplement missing low-frequency components for CNNs and reduce global information loss. Despite its effectiveness, this method utilizes a fixed compensation strategy and thus fails to accommodate diverse feature demands across multi-step forecasting tasks. In ultra-short-term irradiance prediction, short-range forecasting prioritizes high-frequency local features to trace rapid cloud movements and localized solar shading changes. By contrast, long-range forecasting heavily relies on low-frequency global information to depict the overall evolving trends of cloud systems. A static fusion scheme for high-frequency and low-frequency features cannot satisfy the differentiated modeling needs for varied forecasting steps, which restricts the model’s stability and prediction accuracy in multi-step photovoltaic forecasting.

To tackle the aforementioned limitations, this study develops a multimodal fusion model termed IST for ultra-short-term solar irradiance forecasting. The proposed model incorporates three core components: InceptionNeXt, a step-adaptive low-frequency compensation unit, and TempAttnLSTM. The InceptionNeXt network \cite{yu2024inceptionnext} is adopted to extract multi-scale spatial features from ground-based cloud images. Then,  a Step-Adaptive Low-Frequency Compensation Unit (SALFCU) is designed, which introduces step-related conditional information and learnable gating modules. Such design enables the model to dynamically balance high-frequency local details and low-frequency global information to fit the feature requirements of different forecasting steps. Ultimately, the cloud features optimized by SALFCU are fused with meteorological time-series data and imported into the TempAttnLSTM module to model global temporal dependencies and realize multi-step irradiance forecasting.

The main contributions of this paper are summarized as follows:
\begin{enumerate}[label=\alph*)]
\item This work presents an IST-based multimodal fusion framework for ultra-short-term solar irradiance forecasting. By jointly modeling cloud-derived spatial features and meteorological time-series information, the proposed method effectively improves the modeling accuracy for complex cloud movements, time-varying shading and abrupt irradiance variations.

\item The InceptionNeXt architecture is utilized to perform feature extraction on ground-based cloud images. Equipped with a multi-branch deep convolutional design, this network adaptively captures cloud features across multiple scales and orientations. It effectively addresses the inherent drawbacks of conventional single-convolution models, which are incapable of comprehensively characterizing the complex geometric patterns of diverse clouds.

\item A novel Step-Adaptive Low-Frequency Compensation Unit (SALFCU) is elaborately designed in this study.  A step-aware dynamic gating strategy is embedded into the module to adaptively regulate the weights of high-frequency and low-frequency features across diverse forecasting steps. Specifically, the model prioritizes high-frequency local details to capture fast-changing cloud dynamics at earlier prediction steps, while reinforcing low-frequency global trend learning at farther prediction steps. 

\end{enumerate}

The remainder of this paper is organized as follows. Section \ref{sec2} elaborates on the overall framework and key modules of the proposed IST model. Section \ref{sec:experiment} presents the experimental settings, comparative models and ablation.  Section \ref{sec:conclusion} concludes this paper and discusses future research directions.

\section{Methodology}\label{sec2}

\subsection{The overall procedure}\label{sec:overall}

\begin{figure*}[tp]
    \centering
    \includegraphics[width=0.9\linewidth]{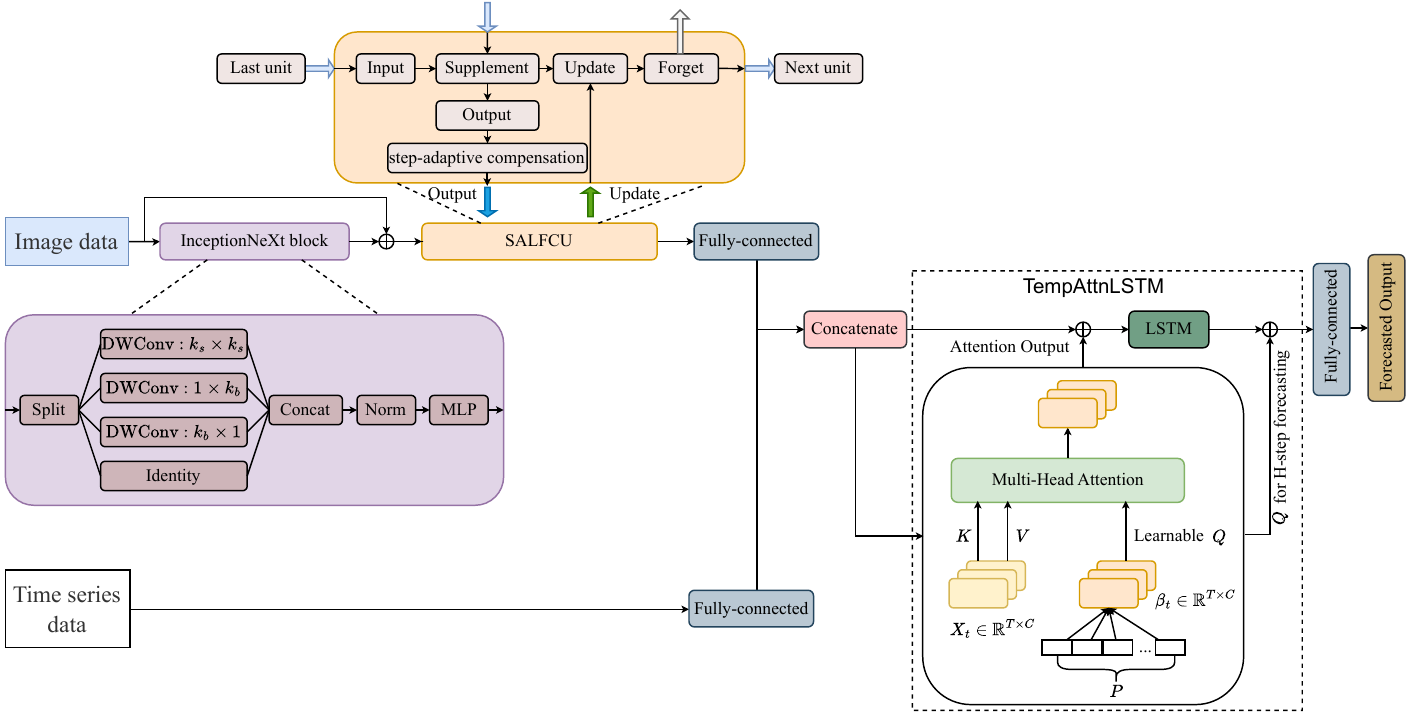}
    \caption{The IST model}\label{fig:ist_framework}
\end{figure*}

A multimodal forecasting model dubbed IST is developed in this work, embedding multi-scale cloud image feature extraction, step-adaptive low-frequency compensation and temporal attention modules.
This model efficiently extracts multi-dimensional visual features from cloud images, covering cloud morphology, spatial distribution, motion trajectories and shading variations. The integration of spatial visual and time-series numerical features empowers the model to capture sudden, non-stationary changes in irradiance amid complicated weather scenarios, which facilitates precise ultra-short-term multi-step solar irradiance forecasting. 
The overall architecture of the IST model is illustrated in Figure \ref{fig:ist_framework}.

To implement multi-scale spatial feature extraction, ground-based cloud image sequences are delivered to the InceptionNeXt network \cite{yu2024inceptionnext}. Built upon a multi-branch parallel convolutional paradigm, this network integrates four complementary branches, including $3 \times 3$ square depthwise separable convolution, $1 \times 11$ horizontal convolution, $11 \times 1$ vertical convolution, and identity mapping. Each branch specializes in distinct feature patterns: a) square convolution extracts local textural and edge details; b) strip convolutions perceive horizontal extensions and vertical variations of clouds; c)identity mapping reserves primitive feature data. These multi-branch features are concatenated across channels and refined through a multi-layer perceptron to generate high-dimensional spatial features. Such a well-designed parallel structure enables adaptive modeling of multi-scale and anisotropic cloud characteristics, remedying the deficiency of single-scale convolution in describing sophisticated cloud geometric features. Details can refer to Section \ref{sec:InceptionNeXt}.

After feature extraction by InceptionNeXt, derived cloud representations are forwarded to the SALFCU. Combining a multi-scale low-frequency memory mechanism and multi-gate collaborative architecture, SALFCU builds a parallel compensation branch to sustain global low-frequency structural characteristics of cloud images. To accommodate varying prediction ranges, a step-aware dynamic gating module is integrated to assign adaptive weights to low-frequency features according to forecasting steps. Such a design allows the model to flexibly optimize feature representations. It concentrates on fine-grained cloud details for short-step prediction and highlights long-term cloud evolution rules for long-step prediction, effectively enhancing multi-step prediction robustness. Detailed procedure can be found in Section \ref{sec:SALFCU}.

Eventually, the multimodal features consisting of meteorological attributes and refined cloud representations are fused and delivered to the TempAttnLSTM module to implement global temporal learning and multi-step prediction. As a specialized temporal attention unit tailored for ultra-short-term irradiance prediction \cite{wang2025multi}, TempAttnLSTM augments conventional LSTM with multi-head attention, learnable periodic queries and future time-oriented vectors. This architecture enables the module to perceive long-range temporal dependencies, periodic variations and cross-step correlations from time-series samples. In the proposed IST framework, the module explores inherent global temporal characteristics of multimodal fusion features and generates targeted multi-step outputs under the guidance of future time queries. A fully connected layer is adopted to derive the final $H$-step solar irradiance forecasting sequence. Specific information is presented in Section \ref{sec:TempAttnLSTM}.

\subsection{InceptionNeXt}\label{sec:InceptionNeXt}

Ground-based sky images contain cloud structures with inherent multi-scale and directional heterogeneity, which vary significantly in scale, boundary shape, extension orientation, and motion patterns. InceptionNeXt \cite{yu2024inceptionnext} is introduced for sky image feature extraction. Its parallel multi-branch depthwise convolution design enables the joint acquisition of local texture details and directional structural information at a manageable computational cost.

For each time step, the extracted ground-based sky image feature is defined as \(I\in \mathbb{R}^{H_I \times W_I \times C_I}\), with \(H_I\), \(W_I\), and \(C_I\) referring to its height, width and channel count. InceptionNeXt initially divides the feature \(I\) into four channel-level branch groups, as follows:
\begin{equation}
	\begin{split}
		I_{\rm hw},\ I_{\rm w},\ I_{\rm h},\ I_{\rm id} 
		&= \text{Split}(I) \\
		&= I_{:,:,g},\ I_{:,g:2g},\ I_{:,2g:3g},\ I_{:,3g:},
	\end{split}
	\label{eq:split}
\end{equation}
where \(I_{\rm hw}\), \(I_{\rm w}\), \(I_{\rm h}\), and \(I_{\rm id}\) are fed into the \(k_s \times k_s\) square convolution branch, \(1 \times k_b\) horizontal band convolution branch, \(k_b \times 1\) vertical band convolution branch, and identity mapping branch, respectively. The grouped channel number is defined as \(g=r_g C\), where \(r_g\) is the channel allocation ratio. In Eq.~\eqref{eq:split}, \(I_{:,:,g}\), \(I_{:,g:2g}\), \(I_{:,2g:3g}\), and \(I_{:,3g:}\) denote four consecutive channel slices.

Every branch employs depthwise convolution to execute channel-independent spatial convolution, thereby reducing the overall parameter quantity and computational overhead. The mathematical formulations of the four branch operations are presented below:
\begin{equation}
	\begin{aligned}
		I'_{\rm hw} &= \text{DWConv}_{k_s \times k_s}^{g \to g}(I_{\rm hw}), \\
		I'_{\rm w}  &= \text{DWConv}_{1 \times k_b}^{g \to g}(I_{\rm w}), \\
		I'_{\rm h}  &= \text{DWConv}_{k_b \times 1}^{g \to g}(I_{\rm h}), \\
		I'_{\rm id} &= I_{\rm id},
	\end{aligned}
	\label{eq:branches}
\end{equation}
where \(k_s\) and \(k_b\) refer to the square kernel size and band kernel size with default values of \(3\) and \(11\), respectively. \(I'_{\rm hw}\), \(I'_{\rm w}\), \(I'_{\rm h}\), and \(I'_{\rm id}\) correspond to the outputs of different branches. \(\text{DWConv}_{k_s \times k_s}\) acquires local cloud textures and fine edge details. \(\text{DWConv}_{1 \times k_b}\) and \(\text{DWConv}_{k_b \times 1}\) separately extract horizontal extended structures and vertical variation features of clouds. The identity branch is designed to reserve the original input channel features.

The four branch outputs are concatenated in the channel dimension to achieve multi-scale feature fusion for sky images:
\begin{equation}
	I' = \text{Concat}(I'_{\rm hw},\ I'_{\rm w},\ I'_{\rm h},\ I'_{\rm id}).
	\label{eq:concat}
\end{equation}
where \(I'\) denotes the multi-scale sky-image feature extracted by InceptionNeXt.

The proposed structure exhibits prominent superiority in feature extraction. The square branch acquires fine-grained local textures and edge features via a small receptive field, matching the characteristics of scattered small clouds and local shadow boundaries. Meanwhile, horizontal and vertical band branches achieve directional receptive field expansion with negligible computational overhead, enabling the effective extraction of large-scale stratiform clouds, band-shaped cloud systems, and directionally moving cloud features. The identity branch further reserves primitive feature information to avoid information degradation from redundant transformation operations.

In ground-based sky image analysis, InceptionNeXt leverages multi-branch depthwise convolution to strengthen the modeling of multi-scale and multi-directional cloud structures. The extracted integrated spatial features are fed into the follow-up step-adaptive low-frequency compensation module. This architecture overcomes the inherent limitation of single-scale CNNs in adapting to complex cloud morphologies, laying a solid foundation for multimodal fusion and accurate multi-step irradiance prediction.

\subsection{SALFCU }\label{sec:SALFCU}

For sky-image-driven ultra-short-term irradiance forecasting, high-frequency spatial details are dominated by local cloud boundaries, fragmented cloud textures and shading changes, while low-frequency global structures are embodied in large-scale cloud distributions, stratiform coverage and overall motion trends. However, hierarchical convolution and downsampling in conventional CNNs inevitably weaken global low-frequency information, thereby undermining the modeling performance for long-term cloud evolution.

To overcome the aforementioned drawback, a Step-Adaptive Low-Frequency Compensation Unit (SALFCU) is developed in this study. With the InceptionNeXt-extracted multi-scale sky image feature \(I'\) as the primary input, the SALFCU establishes a parallel low-frequency memory branch. It integrates six core components, including input, supplement, update, forget, and output gates, as well as a step-adaptive compensation module. These modules collectively propagate prior low-frequency states, mine supplementary low-frequency cues from raw sky images, and fuse backbone and memory feature representations. Through discrete wavelet transform, the model retains valid low-frequency components, produces compensatory features, and dynamically calibrates the compensation magnitude for varying prediction steps.

The input gate transmits prior low-frequency memory features to the current computational unit:
\begin{equation}
	F_{\rm in} = \operatorname{ReLU}\Bigl(\operatorname{BN}\bigl(\operatorname{Conv}_{1\times1}(F_{\rm last})\bigr)\Bigr),
	\label{eq:inputgate}
\end{equation}
where \(F_{\rm last}\) denotes the low-frequency feature from the previous memory unit; \(\operatorname{Conv}_{1\times1}\) adjusts channels and enables cross-channel interaction; \(\operatorname{BN}\) stabilizes feature distributions; \(\operatorname{ReLU}\) is the nonlinear activation function; and \(F_{\rm in}\) is the input-gate output.

The supplement gate extracts additional low-frequency features from the original sky-image information:
\begin{equation}
F_{\rm supp} = \operatorname{ReLU}\Bigl(\operatorname{BN}\bigl(\operatorname{Conv}_{3\times3}(F_{\rm ori})\bigr)\Bigr),
\label{eq}
\end{equation}
where \(F_{\rm ori}\) denotes the original ground-based sky-image feature aligned with the current feature scale, \(\operatorname{Conv}_{3\times3}\) extracts local spatial structures, and \(F_{\rm supp}\) is the supplement-gate output. In implementation, \(F_{\rm ori}\) is aligned with the current memory pathway in both spatial size and channel dimension to ensure valid fusion.

The update gate fuses the backbone feature \(I'\) with \(F_{\rm in}\):
\begin{equation}
	F_{\rm fused} = \operatorname{Concat}(I', F_{\rm in}),
	\label{eq:updategate}
\end{equation}
where \(\operatorname{Concat}\) denotes channel-wise concatenation, and \(F_{\rm fused}\) is used for subsequent low-frequency extraction.

The forget gate decomposes \(F_{\rm fused}\) in the frequency domain using discrete wavelet transform:
\begin{equation}
	F_{\rm memory},\ Y_h = \operatorname{DWT}(F_{\rm fused}),
	\label{eq:forgetgate}
\end{equation}
where \(\operatorname{DWT}\) denotes discrete wavelet transform; \(F_{\rm memory}\) is the retained low-frequency feature and serves as the memory state; and \(Y_h\) denotes the high-frequency detail coefficients. The high-frequency components are not discarded but weakened in the low-frequency pathway to enhance the modeling of macroscopic cloud structures and evolution trends.

The output gate combines \(F_{\rm in}\) and \(F_{\rm supp}\) to generate the low-frequency compensation feature:
\begin{equation}
	F_{\rm out} = F_{\rm in} + F_{\rm supp}.
	\label{eq:outputgate}
\end{equation}
where \(F_{\rm out}\) denotes the output-gate compensation feature. This operation integrates historical low-frequency memory with current supplementary low-frequency information.

SALFCU further introduces a prediction-step-aware adaptive compensation mechanism:
\begin{equation}
	F_{\rm final} = \alpha(H) \cdot F_{\rm out},
	\label{eq:adaptive}
\end{equation}
where \(H\) denotes the current prediction step; \(\alpha(H)\) is the adaptive weighting coefficient determined by \(H\); and \(F_{\rm final}\) is the step-adapted low-frequency compensation feature. This design assigns step-dependent weights to low-frequency information, improving feature adaptability in multi-step forecasting.

The adaptive weighting coefficient \(\alpha(H)\) is generated by a learnable embedding layer and a fully connected mapping:
\begin{equation}
	\alpha(H) = \sigma(W_g \cdot \operatorname{Embed}(H) + b_g),
	\label{eq:alpha}
\end{equation}
where \(\operatorname{Embed}(H)\) maps the discrete prediction step into a continuous vector space; \(W_g\) is the learnable weight matrix; \(b_g\) is the learnable bias vector; and \(\sigma\) is the sigmoid activation function. SALFCU abandons static fixed compensation schemes and dynamically learns the correlation between prediction step and compensation demand in the training phase. It targets abrupt local cloud changes at earlier prediction steps and focuses on global low-frequency structural features at farther prediction steps, thereby precisely modeling cloud evolution patterns.

\subsection{TempAttnLSTM}\label{sec:TempAttnLSTM}

To effectively model global temporal dependencies of multimodal fused features, the TempAttnLSTM module is employed in this paper \cite{wang2025multi}. This module integrates a temporal attention mechanism with learnable periodic queries before standard LSTM temporal modeling, which enhances the modeling capacity for temporal periodicity, variable interactions, and long-range dependencies. By feeding the fused sky image and meteorological time-series features into TempAttnLSTM, high-quality temporal representations can be generated to support accurate multi-step irradiance forecasting.

Practical irradiance data commonly suffer from local noise, outliers, and transient meteorological disturbances. Directly deriving query vectors from raw sequences increases model sensitivity to such interference, impairing global temporal dependency modeling. Accordingly, TempAttnLSTM employs a learnable query matrix to mine global variable correlations, where the query matrix \(Q\) is formulated as a periodic learnable matrix \(\beta_t \in \mathbb{R}^{T \times C}\):
\begin{equation}
	Q = \beta_t \in \mathbb{R}^{T \times C}
	\label{eq:query_matrix}
\end{equation}
where \(Q\) is the query matrix in temporal attention; \(\beta_t\) denotes the periodic learnable query matrix at time \(t\); \(T\) is the input sequence length; and \(C\) is the feature dimension.

The matrix \(\beta_t\) follows a periodic pattern with period \(P\):
\begin{equation}
	\beta_t = \beta_{(t + i \cdot P)}
	\label{eq:periodic_beta}
\end{equation}
where \(P\) is the predefined period length, \(i\) is an integer index, and \(\beta_{(t+i\cdot P)}\) denotes the query representation shifted by \(i\) periods from time \(t\). This design introduces temporal priors into sequence representation, improving the modeling of long-term dependencies and recurrent patterns.

In multi-head attention, the output of each attention head \(\text{head}_h\) is computed as
\begin{equation}
	\text{head}_h = \text{Softmax}\left( \frac{Q_h K_h^\top}{\sqrt{T}} \right) V_h
	\label{eq:head_h}
\end{equation}
where \(\text{head}_h\) denotes the output of the \(h\)-th attention head, \(Q_h = \beta_t W_h^Q\), \(K_h = X_t W_h^K\), and \(V_h = X_t W_h^V\). Here, \(Q_h\), \(K_h\), and \(V_h\) are the query, key, and value matrices of the \(h\)-th attention head, respectively; \(W_h^Q\), \(W_h^K\), and \(W_h^V\) are the corresponding learnable projection matrices; \(X_t\) is the TempAttn input sequence, i.e., the multimodal fused temporal feature; \(\text{Softmax}(\cdot)\) normalizes attention weights; and \(\sqrt{T}\) scales attention scores to alleviate numerical instability caused by increasing sequence length.

The multi-head attention output is obtained by concatenating all attention heads and applying a linear projection with \(W^O\):
\begin{equation}
	\text{MHA}(Q, K, V) = \text{Concat}(\text{head}_1, \dots, \text{head}_h) W^O
	\label{eq:mha}
\end{equation}
where \(\text{MHA}(Q,K,V)\) denotes the multi-head attention output, \(\text{Concat}(\cdot)\) denotes feature-wise concatenation, and \(W^O\) is a learnable output projection matrix that maps the concatenated multi-head features into the target feature space.

This design empowers TempAttn to incorporate global periodic priors from \(\beta_t\) and preserve local input details via key-value pairs derived from \(X_t\), improving the capability of capturing multivariate temporal dependencies. Within the TempAttnLSTM framework, residual connections integrate TempAttn-enhanced temporal features with original sequential inputs, and the fused features are delivered to LSTM for dynamic temporal modeling. Guided by future \(H\)-step query vectors, the LSTM learns step-aware high-level temporal representations tailored for different prediction steps. Consequently, TempAttnLSTM synergizes global periodic characteristics, local dynamic recurrence, and future-step prior information to provide reliable temporal feature support for multi-step irradiance forecasting.

\begin{algorithm}[!tp]
	\caption{IST multimodal fusion algorithm for ultra-short-term PV irradiance forecasting}
	\label{alg:IST}
	\begin{algorithmic}[1]
		\Require Ground-based sky image sequence \(I\), meteorological time-series data \(X\), prediction step \(H\), small square kernel size \(k_s\), and band kernel size \(k_b\)
		\Ensure Future \(H\)-step PV irradiance forecasts
		
		\Statex \textbf{Phase I. InceptionNeXt-based multi-scale sky image feature extraction}
		\State Split \(I\) into four channel-wise branch groups.
		\State Apply depthwise separable convolutions with kernels \(k_s \times k_s\), \(1 \times k_b\), and \(k_b \times 1\), while preserving one identity branch.
		\State Concatenate all branch outputs along the channel dimension to obtain \(I'\).
		
		\Statex \textbf{Phase II. SALFCU-based step-adaptive low-frequency compensation}
		\State Feed \(I'\) into the SALFCU module.
		\For{each SALFCU layer}
			\State Generate \(F_{\rm in}\) by transferring the previous low-frequency memory feature.
			\State Generate \(F_{\rm supp}\) by injecting multi-scale low-frequency information from the original sky image.
			\State Concatenate \(I'\) and \(F_{\rm in}\) to obtain \(F_{\rm fused}\).
			\State Apply discrete wavelet transform to \(F_{\rm fused}\) to suppress high-frequency details and retain \(F_{\rm memory}\).
			\State Compute \(F_{\rm out}=F_{\rm in}+F_{\rm supp}\).
			\State Generate \(\alpha(H)\) from \(H\) using a learnable embedding layer and fully connected layers.
			\State Compute \(F_{\rm final} \gets \alpha(H) \cdot F_{\rm out}\).
		\EndFor
		
		\Statex \textbf{Phase III. Multimodal feature fusion and TempAttnLSTM-based temporal forecasting}
		\State Standardize \(X\) and project it through a fully connected layer.
		\State Concatenate \(F_{\rm final}\) with the projected temporal feature to form the multimodal fused feature.
		\State Feed the fused feature into TempAttnLSTM for temporal modeling with multi-head attention and a learnable periodic query matrix.
		\State Introduce the future temporal query vector \(Q\) to guide multi-step forecasting.
		\State Generate the future \(H\)-step PV irradiance forecasts through a fully connected layer.
	\end{algorithmic}
\end{algorithm}

\subsection{Summary}

Algorithm~\ref{alg:IST} illustrates the comprehensive forecasting procedure of the IST model. The framework comprises three core modules for sequential forecasting. InceptionNeXt serves as the spatial encoder to capture multi-scale structural cloud features, and efficiently expands spatial receptive fields to characterize diverse cloud morphologies at a low computational cost.. SALFCU performs step-aware low-frequency global compensation and multimodal feature fusion with meteorological time series.  SALFCU leverages a low-frequency memory pathway and step-varying weight \(\alpha(H)\) to preserve global cloud evolution trends and dynamically balance local and global feature representations across different forecasting steps. 
TempAttnLSTM synergizes periodic attention and recurrent LSTM modeling to jointly capture long-term periodicity and local temporal dynamics, to yield multi-step irradiance predictions.
Benefiting from the seamless integration of spatial extraction, frequency compensation, and temporal modeling, IST achieves stable and reliable multi-step irradiance forecasting in complex cloud scenarios.

\section{Experiments and analysis} \label{sec:experiment}

We conduct experiments based on two data sources: the publicly available NREL dataset and an in-situ dataset collected from multiple photovoltaic plants in Shandong. The utilized data cover $3\times 64 \times 64$-resolution ground-based cloud images and synchronous meteorological time-series records. To eliminate experimental biases and achieve equitable model comparison, all models are tested under consistent preprocessing strategies, data splitting rules, hyperparameter configurations and evaluation criteria. The entire experimental campaign is executed on a high-performance server featuring an Intel Xeon Gold 6148 CPU, 256 GB RAM, and an RTX 3090 graphics card with 24 GB VRAM.

\subsection {Comparative methods and evaluation indexes}

To comprehensively verify the performance of the IST model in multimodal ultra-short-term solar irradiance forecasting, nine representative baseline models are selected for comparative experiments. According to input types, these baselines are divided into three categories: single time-series models, single cloud image models and multimodal fusion models.
\begin{enumerate}[label=\alph*)]
\item Single time-series baseline models: Tri-TimesNet-TempAttnLSTM \cite{wang2025multi}, BiLSTM \cite{schuster1997bidirectional}, iTransformer \cite{liu2023itransformer} and PatchTST \cite{nie2022time}. Tri-TimesNet-TempAttnLSTM is embedded with multi-stream decomposition and temporal attention to enhance temporal feature learning. BiLSTM is capable of synchronously extracting forward and backward temporal correlations from meteorological data. iTransformer enhances long-range prediction capability via temporal dimension inversion and attention operations. Differently, PatchTST segments time-series signals into sequential patch tokens under a channel-isolated Transformer paradigm to excavate distant temporal patterns.

\item Single cloud image baseline models: 3DCNN and CNN-ConvLSTM. 3DCNN exploits 3D convolutional operations to jointly learn spatio-temporal representations and perceive evolving cloud patterns within continuous image samples \cite{zhao20193d}. By contrast, CNN-ConvLSTM adoptes a dual-stage learning paradigm. CNN is responsible for single-frame spatial feature extraction, while ConvLSTM captures temporal dependencies among successive cloud images.

\item Multimodal baseline models: MICNN-L \cite{ajith2023deep}, CNN-L \cite{ajith2021deep} and EL \cite{shan2022ensemble}. 
MICNN-L couples a multi-input CNN backbone for visual feature extraction with LSTM units for sequential feature learning. CNN-L achieves deep cross-modal information interaction by fusing cloud imagery and time-series features at the early stage of the LSTM workflow.  EL  leverages the ensemble learning paradigm to assemble predictions yielded by multiple sub-models, thereby improving model robustness and generalization for irradiance forecasting.

\end{enumerate}

Five evaluation metrics are adopted in this paper, including Root Mean Square Error (RMSE), Mean Absolute Error (MAE), normalized RMSE (nRMSE), normalized MAE (nMAE) and coefficient of determination $R^2$. The definitions of these metrics are presented as follows:
\begin{equation}
	\text{RMSE} = \sqrt{\frac{1}{N} \sum_{i=1}^{N} (p_i - a_i)^2}
	\label{eq:rmse}
\end{equation}
\begin{equation}
	\text{MAE} = \frac{1}{N} \sum_{i=1}^{N} |p_i - a_i|
	\label{eq:mae}
\end{equation}
\begin{equation}
	\text{nRMSE} = \dfrac{ \sqrt{ \dfrac{1}{N} \sum_{i=1}^{N} (p_i - a_i)^2 } }{ \bar{a} } \times 100\%
	\label{eq:nrmse}
\end{equation}
\begin{equation}
	\text{nMAE} = \dfrac{ \dfrac{1}{N} \sum_{i=1}^{N} |p_i - a_i| }{ \bar{a} } \times 100\%
	\label{eq:nmae}
\end{equation}
\begin{equation}
	R^2 = 1 - \dfrac{ \sum_{i=1}^{N} (p_i - a_i)^2 }{ \sum_{i=1}^{N} (a_i - \bar{a})^2 }
	\label{eq:r2}
\end{equation}
where $p_i$ denotes the predicted value, $a_i$ represents the corresponding observed value, $N$ is the total number of predicted-observed pairs for evaluation, and $\bar a$ stands for the mean of all observed values.

\subsection{NREL Data}

All experiments are conducted on the benchmark NREL dataset for model evaluation \cite{sengupta2018national}. Spanning 2021–2023, the dataset adopts a 10-minute sampling frequency and incorporates comprehensive variables covering GHI, DNI, DHI, and essential meteorological factors (cloud cover, wind speed, solar azimuth angle). Data preprocessing consists of three sequential steps. To secure valid and continuous data distribution, this paper first filters out nighttime samples and invalid entries with persistent zero GHI within three adjacent time steps. Timestamp-based alignment is subsequently implemented to match multimodal time-series data and cloud images. On this basis, numerical features undergo normalization, and cloud images are unified in resolution and standardized channel-wise. The finalized dataset is partitioned into three subsets with a 4:1:1 ratio for training, validation and testing.

\begin{table*}[!tp]
	\centering
	\caption{Forecasting performance on the NREL data (0-4 hours)}\label{table:nrel}
	\resizebox{1\textwidth}{!}{
		\begin{tblr}{
				hline{1-2,7} = {-}{},
			}
			Metric  & Proposed & {Tri-TimesNet-\\TempAttnLSTM}    & BiLSTM & iTransformer & PatchTST & 3DCNN & {CNN-\\ConvLSTM} & MICNN-L & CNN-L & EL  \\
			MAE   &\textbf{ 91.4520}  &101.4411& 110.1538    & 114.6690          & 118.5189      & 122.5934   & 119.1169          & 107.1158     & 112.2753   & 107.3624 \\
			RMSE  & \textbf{149.2842} &155.0645& 167.6434    & 172.4550          & 168.2460      & 178.7249   & 176.0519          & 164.2914     & 166.2945   & 165.0716 \\
			nMAE  & \textbf{0.2349}  &0.2606 & 0.2830      & 0.2946            & 0.3045        & 0.3149     & 0.3059            & 0.2751       & 0.2884     & 0.2758   \\
			nRMSE & \textbf{0.3834}  &0.3983 & 0.4308      & 0.4431            & 0.4322        & 0.4591     & 0.4522            & 0.4220       & 0.4271     & 0.4240   \\
			R\textsuperscript{2} & \textbf{0.7618} &0.7430  &0.6996		&0.6821	 &0.6974&0.6585&0.6687&0.7115&0.7044&0.7087
	\end{tblr}}
\end{table*}

\begin{figure*}[htp]
	\centering
	\includegraphics[width=1\linewidth]{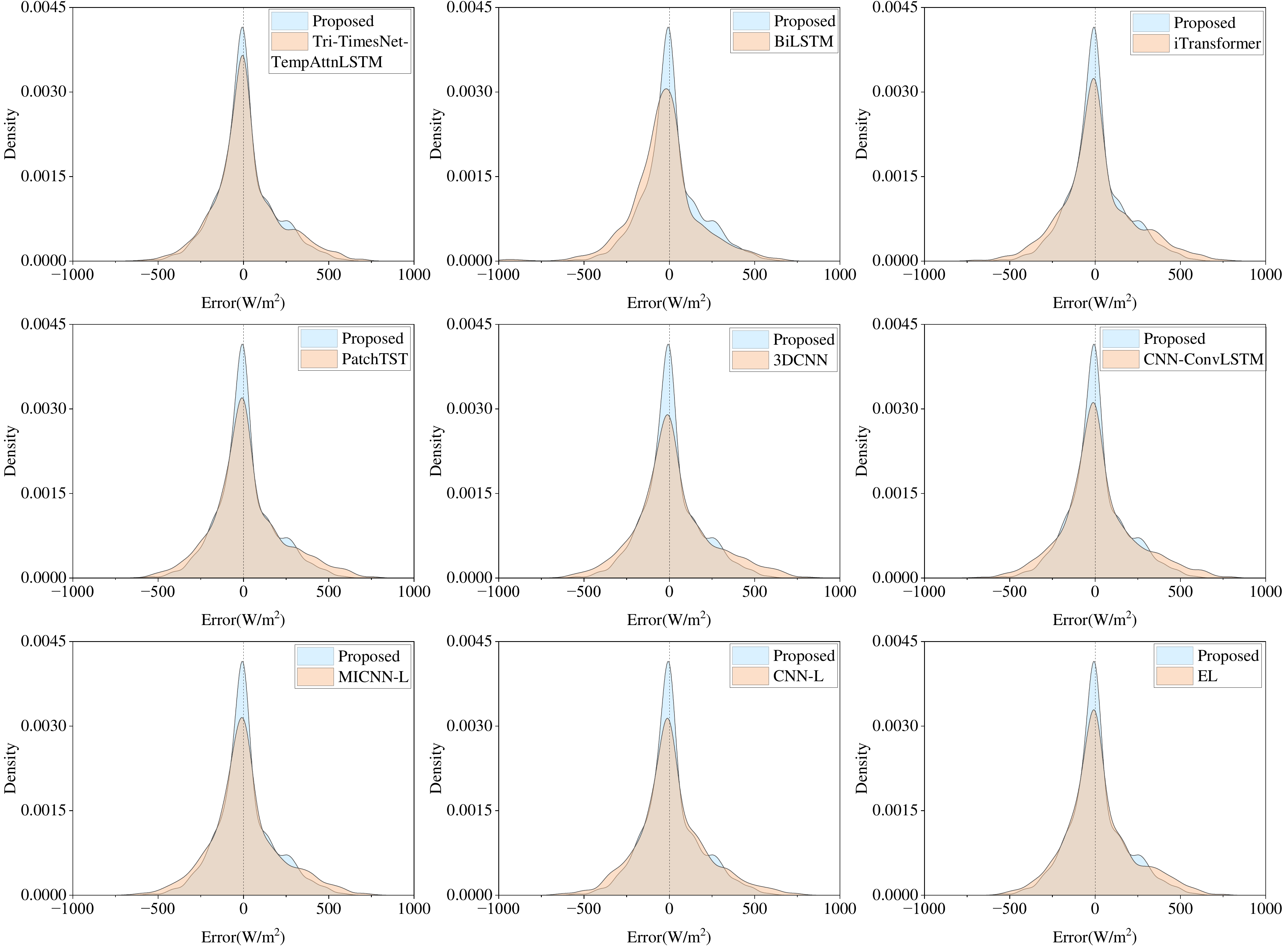} 
	\caption{Error density distribution on the NREL data: the proposed IST model vs. comparative methods}\label{fig:capacity-nrel}
\end{figure*}

\subsubsection{Experimental results and analysis}
\label{subsubsec:nrel_results_analysis}

To validate model effectiveness, comparative experiments are implemented on the NREL dataset with nine baseline models from Section 3.1. The corresponding quantitative results are listed in Table~\ref{table:nrel}. 
It is evident that the proposed IST model achieves the best forecasting performance over the 0--4 h prediction horizon, with the lowest MAE of 91.4520 and RMSE of 149.2842 among all compared methods. The coefficient of determination $R^2$  of the proposed IST model reaches 0.7618, outperforming all baseline models.

The proposed IST achieves superior quantitative performance over all single time-series baselines (Tri-TimesNet-TempAttnLSTM, BiLSTM, iTransformer, and PatchTST) in terms of MAE and RMSE, with respective MAE reductions of 9.8\%, 17.0\%, 20.2\%, 22.8\% and RMSE reductions of 3.7\%, 10.9\%, 13.4\%, 11.3\%. Although these models possess powerful temporal modeling capabilities for meteorological sequences, their single-modal input design excludes spatial cloud information, making them incapable of capturing abrupt irradiance variations caused by dynamic cloud shading. In contrast, the multimodal design of IST fuses cloud spatial features and meteorological temporal features, greatly strengthening the model’s ability to perceive cloud-induced irradiance fluctuations.

\begin{table*}[!tp]
	\centering
	\caption{Ablation study on the NREL data}\label{table_ablation_nrel}
		\begin{tblr}{
				cell{2}{2} = {font=\bfseries},
				cell{3}{2} = {font=\bfseries},
				cell{4}{2} = {font=\bfseries},
				cell{5}{2} = {font=\bfseries},
				cell{6}{2} = {font=\bfseries},
				hline{1-2,7} = {-}{},
			}
			Metric  & Proposed     & w/o Image & w/o Time Series & w/o InceptionNeXt & w/o SALFCU & w/o TempAttn \\
			MAE   & 91.4520  & 108.1538  & 117.3021        & 98.5189           & 102.5934   & 103.1169     \\
			RMSE  & 149.2842 & 165.6434  & 174.0461        & 156.7546          & 162.7249   & 161.0716     \\
			nMAE  & 0.2349   & 0.2778    & 0.3012          & 0.2531            & 0.2635     & 0.2649       \\
			nRMSE & 0.3834   & 0.4254    & 0.4470         & 0.4027            & 0.4180     & 0.4138   \\
			R\textsuperscript{2} &0.7618  &0.7067  &0.6821  &0.7373 &0.7169   	&0.7227	 		    
	\end{tblr}
\end{table*}

IST also surpasses two typical single-image models (3DCNN and CNN-ConvLSTM), yielding 25.4\%/23.2\% MAE reductions and 16.5\%/15.2\% RMSE reductions. Pure image-based methods can only extract spatial cloud characteristics but lack long-term meteorological temporal cues, resulting in insufficient modeling of overall irradiance evolution. Additionally, conventional convolutional networks suffer from inherent low-frequency information loss due to local receptive field constraints, degrading long-step prediction accuracy. IST remedies this issue by combining multi-scale spatial extraction (InceptionNeXt) and step-adaptive frequency compensation (SALFCU), which dynamically balances high-frequency cloud details and low-frequency global trends to adapt to short- and long-term forecasting tasks.

Compared with state-of-the-art multimodal methods (MICNN-L, CNN-L, EL), IST achieves prominent performance improvements with 14.6\%, 18.5\%, 14.8\% MAE reductions and 9.1\%, 10.2\%, 9.6\% RMSE reductions. Existing multimodal models fail to fully extract multi-scale cloud structural features and lack adaptive frequency adjustment mechanisms for varying prediction steps, restricting their ability to track complex cloud evolution patterns. Benefiting from the organic integration of spatial extraction, frequency compensation, and step-aware temporal modeling, IST realizes effective multimodal feature fusion and step-adaptive prediction optimization. The above results collectively demonstrate the outstanding performance, robustness, and generalization of the proposed model for ultra-short-term irradiance forecasting.

The error density distributions of all models are visualized in Figure \ref{fig:capacity-nrel}, which validates the effectiveness of the multimodal fusion design of IST. The proposed model yields a unimodal error distribution with a higher zero-centered peak and narrower tails. Compared with Tri-TimesNet-TempAttnLSTM, a typical single time-series model, IST produces a more prominent peak and substantially eliminates extreme errors, demonstrating the effectiveness of introducing sky image spatial information for accuracy improvement. By contrast, other baseline models suffer from dispersed error distributions with lower peaks and wider tails, resulting in degraded overall forecasting performance.

\subsubsection{Ablation study}

Ablation experiments are conducted on the NREL dataset to further validate the effectiveness of each core component. All model variants adopt identical experimental configurations and evaluation metrics. Five control groups are designed as follows:
\begin{enumerate}[label=\alph*)]
\item w/o Image: The InceptionNeXt and SALFCU modules are removed, and only TempAttnLSTM is used to process time-series data. This group verifies the necessity of multimodal fusion.
\item w/o Time Series: Time-series inputs are eliminated, and only cloud image sequences are processed under the IST framework to explore the performance gain brought by time-series information.
\item w/o InceptionNeXt: The multi-scale convolutional network is replaced with a standard CNN to verify the effectiveness of the spatial feature extraction structure of InceptionNeXt.
\item w/o SALFCU: The adaptive high- and low-frequency compensation unit is removed to analyze its contribution to balancing frequency information of features.
\item w/o TempAttn: The temporal attention module is replaced with a conventional LSTM to validate the effectiveness of temporal attention and future query vectors.
\end{enumerate}

Ablation comparison experiments are carried out and the results are presented in Table~\ref{table_ablation_nrel}. The full IST model achieves the best overall performance, with an MAE of 91.4520 and an RMSE of 149.2842. Compared with w/o Image, the MAE and RMSE are reduced by 15.4\% and 9.9\%, respectively, indicating that single time-series inputs are insufficient to represent spatial cloud information and abrupt irradiance fluctuations induced by complex cloud shading. Compared with w/o Time Series, which produces the largest error among all variants, the MAE and RMSE are decreased by 20.2\% and 13.4\%, respectively, illustrating that cloud image sequences alone cannot effectively capture the long-term statistical characteristics of meteorological variables or the overall temporal evolution of irradiance. Compared with w/o InceptionNeXt, the full model reduces MAE and RMSE by 7.2\% and 4.8\%, respectively, confirming that the multi-scale and multi-directional convolutional structure is beneficial for extracting discriminative features from complex cloud patterns. 

Compared with w/o SALFCU, the reductions in MAE and RMSE reach 10.9\% and 8.3\%, respectively, verifying that step-adaptive low-frequency compensation contributes to dynamically balancing local high-frequency cloud details and global low-frequency evolution trends across different forecasting steps. Compared with w/o TempAttn, the MAE and RMSE are reduced by 11.3\% and 7.3\%, respectively, which validates the effectiveness of temporal attention and future query vectors in enhancing multi-step temporal dependency modeling. 
Furthermore, the $R^2$  of IST reaches 0.7618, while all ablation variants show declines in  $R^2$  to varying degrees. Specifically, the $R^2$  of w/o Time Series drops to 0.6821, and that of w/o Image decreases to 0.7067. 
These results collectively illustrate that InceptionNeXt, SALFCU and TempAttnLSTM provide complementary improvements in spatial representation, frequency-domain compensation and temporal forecasting, thereby supporting the superior performance of the proposed IST model.

With the effective combination of InceptionNeXt and SALFCU, the IST model realizes dynamic balance of cross-modal information and adapts to complex weather conditions and multi-step forecasting tasks. The experimental results fully validate the effectiveness of the proposed model for irradiance forecasting.

\begin{table*}[!tp]
	\centering
	\caption{Forecasting performance on the practical Shandong PV data (0-4 hours)}\label{result_shandong}
	\resizebox{1\textwidth}{!}{
		\begin{tblr}{
				hline{1-2,7} = {-}{},
			}
			Metric  & Proposed & {Tri-TimesNet-\\TempAttnLSTM}    & BiLSTM & iTransformer & PatchTST & 3DCNN & {CNN-\\ConvLSTM} & MICNN-L & CNN-L & EL  \\
			MAE   & \textbf{50.8559} &58.2525& 83.4586     & 94.8563           & 97.4198       & 96.1642    & 91.6860           & 66.1228      & 70.5490    & 65.7438 \\
			RMSE  & \textbf{68.5293} &79.4482& 111.4231    & 121.8989          & 122.6147      & 120.9179   & 114.1557          & 89.7757      & 93.4905    & 88.1087 \\
			nMAE  & \textbf{0.1980} &0.2307 & 0.3250      & 0.3704            & 0.3768        & 0.3743     & 0.3569            & 0.2573       & 0.2746     & 0.2559  \\
			nRMSE & \textbf{0.2668} &0.3146 & 0.4338      & 0.4746            & 0.4774        & 0.4707     & 0.4444            & 0.3494       & 0.3639     & 0.3429  \\
			R\textsuperscript{2} & \textbf{0.8386} &0.7831	&0.5733	&0.4893	&0.4833	&0.4975	&0.5521	&0.7230	&0.6996	&0.7332		
	\end{tblr}
    }
\end{table*}

\begin{figure*}[!tp]
	\centering
	\includegraphics[width=1\linewidth]{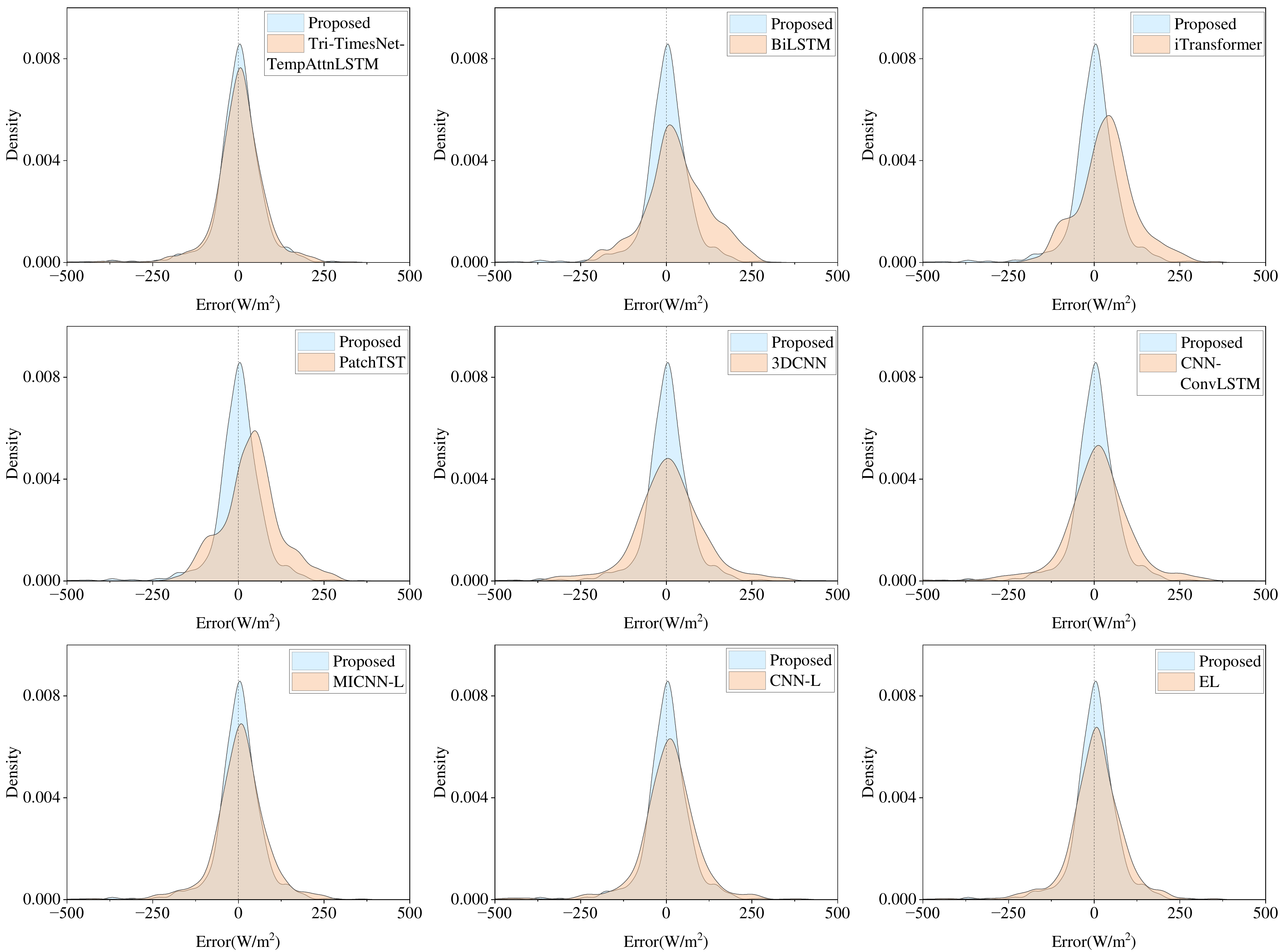} 
	\caption{Error density distribution on the practical Shandong PV data: the proposed IST model vs. comparative methods}
	\label{fig:error-shandong}
\end{figure*}

\subsection{Practical data}

The dataset used in this work is collected from an actual grid-connected photovoltaic power station in Jining, Shandong Province (35.493572° N, 116.280441° E). It has a temporal resolution of 5 minutes and spans from May 21 to December 18, 2024, covering a total period of approximately seven months. The time-series features include key meteorological parameters such as Global Horizontal Irradiance (GHI), temperature, relative humidity, wind speed, wind direction and atmospheric pressure.

Given the limited total number of samples, the dataset is split into training, validation and test sets at a ratio of 8:1:1 to ensure representative and reasonable data distribution across subsets. Uniform preprocessing is performed before model input. We align time-series data and cloud images by timestamps, normalize numerical features, and unify the resolution and standardize channels of cloud images. This approach guarantees consistency of multimodal inputs in both temporal dimension and numerical scale.

\begin{table*}[!tp]
	\centering
	\caption{Ablation study on the practical Shandong PV data}\label{ablation-shandong}
		\begin{tblr}{
				cell{2}{2} = {font=\bfseries},
				cell{3}{2} = {font=\bfseries},
				cell{4}{2} = {font=\bfseries},
				cell{5}{2} = {font=\bfseries},
				cell{6}{2} = {font=\bfseries},
				hline{1-2,7} = {-}{},
			}
			Metric  & Proposed    & w/o Image & w/o Time Series & w/o InceptionNeXt & w/o SALFCU & w/o TempAttn \\
			MAE   & 50.8559 & 69.5534   & 68.3793         & 55.2098           & 54.8434    & 62.0535      \\
			RMSE  & 68.5293 & 88.1758   & 85.1892         & 76.2388           & 77.4621    & 80.0657      \\
			nMAE  & 0.1980  & 0.2707    & 0.2661          & 0.2148            & 0.2134     & 0.2414       \\
			nRMSE & 0.2668  & 0.3432    & 0.3315          & 0.2966            & 0.3014     & 0.3115 \\
			R\textsuperscript{2} &0.8386  &0.7328  &0.7506  &0.8002  &  0.7938	&0.7797	 	 			
	\end{tblr}
\end{table*}

\subsubsection{Experimental results and analysis}
\label{subsubsec:shandong_results_analysis}

The proposed model also achieves remarkable performance on the real-world dataset from the photovoltaic power station in Shandong, and the experimental results are presented in Table \ref{result_shandong}. The MAE of our model is 50.8559, which is reduced by 12.7\%, 39.1\%, 46.4\%, 47.8\%, 47.1\%, 44.5\%, 23.1\%, 27.9\% and 22.7\% compared with Tri-TimesNet-TempAttnLSTM, BiLSTM, iTransformer, PatchTST, 3DCNN, CNN-ConvLSTM, MICNN-L, CNN-L and EL, respectively. Its RMSE reaches 68.5293, representing a reduction of 13.7\%, 38.5\%, 43.8\%, 44.1\%, 43.3\%, 40.0\%, 23.7\%, 26.7\% and 22.2\% against the above nine models in sequence. Besides, the coefficient of determination $R^2$ of the proposed model reaches 0.8386, which is superior to that of all baseline models.

The error density distributions of all models on the practical dataset from the Shandong PV station are visualized in Figure \ref{fig:error-shandong}, which clearly reveals the prominent prediction superiority of IST in real-world scenarios. The proposed model delivers symmetric unimodal error distribution with the sharpest zero-centered peak, minimal dispersion, and narrowest error tails, surpassing all comparative methods in overall prediction accuracy. Compared with the single time-series baseline Tri-TimesNet-TempAttnLSTM, IST better eliminates extreme prediction errors and achieves a more concentrated error distribution. 
 Compared with baseline models, the proposed model not only restrains prediction errors effectively, but also accurately characterizes the inherent variation patterns and long-term evolution trends of irradiance under practical operating conditions.

Compared with the public NREL dataset, the model achieves more substantial error reduction in practical power station scenarios. This indicates that the IST model can well adapt to high-resolution data and complex local meteorological conditions in engineering applications. It effectively mitigates interferences caused by random cloud movements and surrounding terrain of power stations, and possesses excellent practicability, robustness and generalization ability for real engineering scenarios.

\subsubsection{Ablation}

Ablation experiments are conducted on the Shandong photovoltaic dataset, and the results are presented in Table~\ref{ablation-shandong}. The full IST model achieves the optimal performance across all evaluation metrics, with an MAE of 50.8559 and an RMSE of 68.5293. For the variants w/o Image and w/o Time Series, the full model reduces MAE by 26.9\% and 25.6\%, and decreases RMSE by 22.3\% and 19.6\%. These two variants yield the largest increase in prediction errors among all ablation models, indicating that single-modal data cannot accurately characterize the spatiotemporal complexity of cloud evolution in real scenarios, thereby further verifying the necessity of adopting a multimodal data fusion strategy for photovoltaic irradiance prediction. When w/o InceptionNeXt is used, the MAE and RMSE reductions achieved by the full model are 7.9\% and 10.1\%, confirming the importance of multi-scale feature extraction for cloud representation under dynamic real-world cloud conditions. For w/o SALFCU, the full model lowers MAE from this variant by 7.3\% and RMSE by 11.5\%, demonstrating that adaptive high-low frequency compensation plays a vital role in enhancing model performance. Compared with w/o TempAttn, the full model achieves reductions of 18.0\% in MAE and 14.4\% in RMSE, which confirms the effectiveness of the temporal attention mechanism of TempAttnLSTM in real photovoltaic forecasting scenarios. In addition, compared with the NREL dataset, the performance degradation after removing core components is more significant on the real power station dataset, demonstrating that the proposed model is well adapted to high-resolution data in practical engineering and complex, variable local meteorological conditions.

\section{Conclusion}\label{sec:conclusion}

This study develops a novel multimodal fusion IST framework for ultra-short-term solar irradiance forecasting. The proposed model fuses ground-based cloud images and meteorological time-series data to effectively characterize dynamic complex cloud patterns, local shading changes, and non-stationary irradiance variations. It leverages InceptionNeXt for multi-scale and multi-directional spatial feature extraction, SALFCU for step-adaptive low-frequency feature compensation, and TempAttnLSTM for global temporal dependency modeling and multi-step prediction. Experimental validations on the public NREL dataset and practical field data from Shandong photovoltaic stations demonstrate that IST outperforms conventional time-series, image-based, and advanced multimodal baseline models. Ablation tests further confirm the essential contributions of dual-modal inputs and core modules to the model’s superior forecasting capability.

\bibliographystyle{elsarticle-num}
\bibliography{reference}

@article{zhen2020deep,
	title={Deep learning based surface irradiance mapping model for solar {PV} power forecasting using sky image},
	author={Zhen, Zhao and Liu, Jiaming and Zhang, Zhanyao and Wang, Fei and Chai, Hua and Yu, Yili and Lu, Xiaoxing and Wang, Tieqiang and Lin, Yuzhang},
	journal={IEEE Transactions on Industry Applications},
	volume={56},
	number={4},
	pages={3385--3396},
	year={2020},
	publisher={IEEE}
}

@article{feng2022convolutional,
	title={Convolutional neural networks for intra-hour solar forecasting based on sky image sequences},
	author={Feng, Cong and Zhang, Jie and Zhang, Wenqi and Hodge, Bri-Mathias},
	journal={Applied Energy},
	volume={310},
	pages={118438},
	year={2022},
}

@article{huang20233d,
  title={A {3D} {ConvLSTM-CNN} network based on multi-channel color extraction for ultra-short-term solar irradiance forecasting},
  author={Huang, Xiaoqiao and Liu, Jun and Xu, Shaozhen and Li, Chengli and Li, Qiong and Tai, Yonghang},
  journal={Energy},
  volume={272},
  pages={127140},
  year={2023},
  publisher={Elsevier}
}

@article{zang2024improving,
  title={Improving ultra-short-term photovoltaic power forecasting using a novel sky-image-based framework considering spatial-temporal feature interaction},
  author={Zang, Haixiang and Chen, Dianhao and Liu, Jingxuan and Cheng, Lilin and Sun, Guoqiang and Wei, Zhinong},
  journal={Energy},
  volume={293},
  pages={130538},
  year={2024},
  publisher={Elsevier}
}

@article{nie2024sky,
	title={Sky image-based solar forecasting using deep learning with heterogeneous multi-location data: Dataset fusion versus transfer learning},
	author={Nie, Yuhao and Paletta, Quentin and Scott, Andea and Pomares, Luis Martin and Arbod, Guillaume and Sgouridis, Sgouris and Lasenby, Joan and Brandt, Adam},
	journal={Applied Energy},
	volume={369},
	pages={123467},
	year={2024},
	publisher={Elsevier}
}

@article{jonathan2024radiant,
	title={A radiant shift: Attention-embedded {CNNs} for accurate solar irradiance forecasting and prediction from sky images},
	author={Jonathan, Anto Leoba and Cai, Dongsheng and Ukwuoma, Chiagoziem C and Nkou, Nkou Joseph Junior and Huang, Qi and Bamisile, Olusola},
	journal={Renewable Energy},
	volume={234},
	pages={121133},
	year={2024},
	publisher={Elsevier}
}

@article{xu2024minutely,
	title={Minutely multi-step irradiance forecasting based on all-sky images using {LSTM}-InformerStack hybrid model with dual feature enhancement},
	author={Xu, Shaozhen and Liu, Jun and Huang, Xiaoqiao and Li, Chengli and Chen, Zaiqing and Tai, Yonghang},
	journal={Renewable Energy},
	volume={224},
	pages={120135},
	year={2024},
	publisher={Elsevier}
}

@article{dai2023photovoltaic,
	title={Photovoltaic power prediction based on sky images and tokens-to-token vision transformer},
	author={Dai, Qiangsheng and Hou, Xuesong and Su, Dawei and Cui, Zhiwei},
	journal={International Journal of Renewable Energy Development},
	volume={12},
	number={6},
	pages   = {1104--1112},
	year={2023}
}

@article{shi2025ground,
	title={A ground-based cloud image classification method for photovoltaic power prediction based on convolutional neural networks and vision transformer},
	author={Shi, Chaojun and Zhang, Mengyu and Xiang, Hongyin and Zhang, Ke and Ju, Sihao and Zhang, Xiaoyun and Han, Leile},
	journal={Engineering Applications of Artificial Intelligence},
	volume={159},
	pages={111582},
	year={2025},
	publisher={Elsevier}
}

@article{ma2025research,
  title={Research on ultra-short-term photovoltaic power forecasting using multimodal data and ensemble learning},
  author={Ma, Yifeng and Yu, Wenzheng and Zhu, Junyu and You, Zhiyuan and Jia, Aiqing},
  journal={Energy},
  volume={330},
  pages={136831},
  year={2025},
  publisher={Elsevier}
}

@article{caldas2019very,
  title={Very short-term solar irradiance forecast using all-sky imaging and real-time irradiance measurements},
  author={Caldas, M and Alonso-Su{\'a}rez, R},
  journal={Renewable Energy},
  volume={143},
  pages={1643--1658},
  year={2019},
  publisher={Elsevier}
}

@article{ajith2023deep,
	title={Deep learning algorithms for very short term solar irradiance forecasting: A survey},
	author={Ajith, Meenu and Mart{\'\i}nez-Ram{\'o}n, Manel},
	journal={Renewable and Sustainable Energy Reviews},
	volume={182},
	pages={113362},
	year={2023},
	publisher={Elsevier}
}

@article{paletta2023advances,
  title={Advances in solar forecasting: Computer vision with deep learning},
  author={Paletta, Quentin and Terr{\'e}n-Serrano, Guillermo and Nie, Yuhao and Li, Binghui and Bieker, Jacob and Zhang, Wenqi and Dubus, Laurent and Dev, Soumyabrata and Feng, Cong},
  journal={Advances in Applied Energy},
  volume={11},
  pages={100150},
  year={2023},
  publisher={Elsevier}
}

@article{hendrikx2024all,
  title={All sky imaging-based short-term solar irradiance forecasting with Long Short-Term Memory networks},
  author={Hendrikx, NY and Barhmi, K and Visser, LR and De Bruin, TA and P{\'o}, M and Salah, AA and Van Sark, WGJHM},
  journal={Solar Energy},
  volume={272},
  pages={112463},
  year={2024},
  publisher={Elsevier}
}

@article{ansong2025very,
  title={Very short-term solar irradiance forecasting based on open-source low-cost sky imager and hybrid deep-learning techniques},
  author={Ansong, Martin and Huang, Gan and Nyang’onda, Thomas N and Musembi, Robinson J and Richards, Bryce S},
  journal={Solar Energy},
  volume={294},
  pages={113516},
  year={2025},
  publisher={Elsevier}
}

@inproceedings{wu2024multiscale,
	title={Multiscale low-frequency memory network for improved feature extraction in convolutional neural networks},
	author={Wu, Fuzhi and Wu, Jiasong and Kong, Youyong and Yang, Chunfeng and Yang, Guanyu and Shu, Huazhong and Carrault, Guy and Senhadji, Lotfi},
	booktitle={Proceedings of the AAAI Conference on Artificial Intelligence},
	volume={38},
	pages={5967--5975},
	year={2024}
}

@inproceedings{yu2024inceptionnext,
	title={Inceptionnext: When inception meets convnext},
	author={Yu, Weihao and Zhou, Pan and Yan, Shuicheng and Wang, Xinchao},
	booktitle={Proceedings of the IEEE/CVF Conference on Computer Vision and Pattern Recognition},
	pages={5672--5683},
	year={2024}
}

@inproceedings{wang2025multi,
  title={Multi-Stream Decomposition with Temporal Attention for Ultra-Short-Term Photovoltaic Irradiance Forecasting},
  author={Wang, Xiaoqin and Wu, Ji and Wang, Shibo and Zhang, Jingxin},
  booktitle={2025 China Automation Congress (CAC)},
  pages={6413--6420},
  year={2025},
  organization={IEEE}
}

@article{schuster1997bidirectional,
	title={Bidirectional recurrent neural networks},
	author={Schuster, Mike and Paliwal, Kuldip K},
	journal={IEEE Transactions on Signal Processing},
	volume={45},
	number={11},
	pages={2673--2681},
	year={1997},
	publisher={Ieee}
}

@article{liu2023itransformer,
	title={iTransformer: Inverted transformers are effective for time series forecasting},
	author={Liu, Yong and Hu, Tengge and Zhang, Haoran and Wu, Haixu and Wang, Shiyu and Ma, Lintao and Long, Mingsheng},
	journal={arXiv preprint arXiv:2310.06625},
	year={2023}
}

@article{nie2022time,
	title={A time series is worth 64 words: Long-term forecasting with transformers},
	author={Nie, Yuqi and Nguyen, Nam H and Sinthong, Phanwadee and Kalagnanam, Jayant},
	journal={arXiv preprint arXiv:2211.14730},
	year={2022}
}

@article{zhao20193d,
	title={{3D-CNN}-based feature extraction of ground-based cloud images for direct normal irradiance prediction},
	author={Zhao, Xin and Wei, Haikun and Wang, Hai and Zhu, Tingting and Zhang, Kanjian},
	journal={Solar Energy},
	volume={181},
	pages={510--518},
	year={2019},
	publisher={Elsevier}
}

@article{ajith2021deep,
	title={Deep learning based solar radiation micro forecast by fusion of infrared cloud images and radiation data},
	author={Ajith, Meenu and Mart{\'\i}nez-Ram{\'o}n, Manel},
	journal={Applied Energy},
	volume={294},
	pages={117014},
	year={2021},
	publisher={Elsevier}
}

@article{shan2022ensemble,
	title={Ensemble learning based multi-modal intra-hour irradiance forecasting},
	author={Shan, Shuo and Li, Chenxi and Ding, Zhetong and Wang, Yiye and Zhang, Kanjian and Wei, Haikun},
	journal={Energy Conversion and Management},
	volume={270},
	pages={116206},
	year={2022},
	publisher={Elsevier}
}

@article{sengupta2018national,
	title={The national solar radiation data base ({NSRDB})},
	author={Sengupta, Manajit and Xie, Yu and Lopez, Anthony and Habte, Aron and Maclaurin, Galen and Shelby, James},
	journal={Renewable and Sustainable Energy Reviews},
	volume={89},
	pages={51--60},
	year={2018},
	publisher={Elsevier}
}

@article{alonso2015skycamera,
  title={The use of a sky camera for solar radiation estimation based on digital image processing},
  author={Alonso-Montesinos, J and Batlles, FJ},
  journal={Energy},
  volume={90},
  pages={377--386},
  year={2015},
  publisher={Elsevier}
}

@article{shi2024cloudswinnet,
  title={CloudSwinNet: A hybrid {CNN-transformer} framework for ground-based cloud images fine-grained segmentation},
  author={Shi, Chaojun and Su, Zibo and Zhang, Ke and Xie, Xiongbin and Zhang, Xiaoyun},
  journal={Energy},
  volume={309},
  pages={133128},
  year={2024},
  publisher={Elsevier}
}

@article{wang2019photovoltaic,
  title={Photovoltaic power forecasting based on {LSTM}-convolutional network},
  author={Wang, Kejun and Qi, Xiaoxia and Liu, Hongda},
  journal={Energy},
  volume={189},
  pages={116225},
  year={2019},
  publisher={Elsevier}
}

@article{rafati2021high,
  title={High dimensional very short-term solar power forecasting based on a data-driven heuristic method},
  author={Rafati, Amir and Joorabian, Mahmood and Mashhour, Elaheh and Shaker, Hamid Reza},
  journal={Energy},
  volume={219},
  pages={119647},
  year={2021},
  publisher={Elsevier}
}

@article{dou2025multimodal,
  title={A multi-modal deep clustering method for day-ahead solar irradiance forecasting using ground-based cloud imagery and time series data},
  author={Dou, Weijing and Wang, Kai and Shan, Shuo and Chen, Mingyu and Zhang, Kanjian and Wei, Haikun and Sreeram, Victor},
  journal={Energy},
  volume={321},
  pages={135285},
  year={2025},
}

\end{document}